\documentclass[conference]{IEEEtran}
\IEEEoverridecommandlockouts
\usepackage{cite}
\usepackage{amsmath,amssymb,amsfonts}
\usepackage{algorithmic}
\usepackage{graphicx}
\usepackage{textcomp}
\usepackage{xcolor}
\usepackage{censor}

\usepackage{cuted}
\usepackage{booktabs}
\usepackage{tabularx}
\usepackage{hyperref}
\usepackage{listings}
\usepackage[font=footnotesize]{caption} % needed for listing captions
\usepackage{float}
\usepackage{threeparttable}
\usepackage{subfigure}
\usepackage{soul}

\def\BibTeX{{\rm B\kern-.05em{\sc i\kern-.025em b}\kern-.08em
    T\kern-.1667em\lower.7ex\hbox{E}\kern-.125emX}}

\usepackage{tikz}
\usetikzlibrary{calc}

\lstdefinestyle{EE}{%
  frame=top,frame=bottom,
  basicstyle=\footnotesize\normalfont\sffamily,
  stepnumber=1,
  numbersep=10pt,
  tabsize=2,
  extendedchars=true,
  breaklines=true,
  captionpos=t,
  mathescape=true,
  stringstyle=\color{white}\ttfamily,
  showspaces=false,
  showtabs=false,
  xleftmargin=10pt,
  framexleftmargin=10pt,
  framexrightmargin=0pt,
  framexbottommargin=5pt,
  framextopmargin=5pt,
  showstringspaces=false
 }

\DeclareCaptionFormat{listing}{\rule{\dimexpr\columnwidth+0pt\relax}{0.4pt}\par\vskip1pt#1#2#3}
\begin{document}

\title{Tricking AI chips into Simulating the Human Brain: A Detailed Performance Analysis}

\author{
     \IEEEauthorblockN{Lennart\ P.\ L.\ Landsmeer\IEEEauthorrefmark{1}}
     \IEEEauthorblockA{
         \textit{Quantum \& Computer} \\
         \textit{Engineering Department} \\
         \textit{Delft University of Technology}\\
         Delft, The Netherlands \\
         \\
         \textit{Dept. of Neuroscience} \\
         \textit{Erasmus Medical Center}\\
         Rotterdam, The Netherlands \\
         \href{https://orcid.org/0000-0003-0010-7249}{ORCID 0000-0003-0010-7249}
     }
     \and
     \IEEEauthorblockN{Max C.W. Engelen\IEEEauthorrefmark{1}}
     \IEEEauthorblockA{
         \textit{Dept. of Neuroscience} \\
         \textit{Erasmus Medical Center}\\
         Rotterdam, The Netherlands \\
         \& \\
         \textit{Maxeler IoT Labs} \\
         Delft, Netherlands \\
         \href{https://orcid.org/0000-0002-5762-1276}{ORCID 0000-0002-5762-1276}
     }
     \and
     \IEEEauthorblockN{Rene Miedema}
     \IEEEauthorblockA{
         \textit{Quantum \& Computer} \\
         \textit{Engineering Department} \\
         \textit{Delft University of Technology}\\
         Delft, The Netherlands \\
         \& \\
         \textit{Dept. of Neuroscience} \\
         \textit{Erasmus Medical Center}\\
         Rotterdam, The Netherlands \\
         \href{https://orcid.org/0000-0002-0447-1083}{ORCID 0000-0002-0447-1083}
     }
     \and
     \IEEEauthorblockN{Christos Strydis}
     \IEEEauthorblockA{
         \textit{Quantum \& Computer} \\
         \textit{Engineering Department} \\
         \textit{Delft University of Technology}\\
         Delft, The Netherlands \\
         \& \\
         \textit{Dept. of Neuroscience} \\
         \textit{Erasmus Medical Center}\\
         Rotterdam, The Netherlands \\
         \href{https://orcid.org/0000-0002-0935-9322}{ORCID 0000-0002-0935-9322}
     }
}

\maketitle
\thispagestyle{plain}
\pagestyle{plain}

\begin{abstract}
    Challenging the Nvidia monopoly, dedicated AI-accelerator chips have begun emerging for tackling the computational challenge that the inference and, especially, the training of modern deep neural networks (DNNs) poses to modern computers. The field has been ridden with studies assessing the performance of these contestants across various DNN model types. However, AI-experts are aware of the limitations of current DNNs and have been working towards the fourth AI wave which will, arguably, rely on more biologically inspired models, predominantly on spiking neural networks (SNNs). At the same time, GPUs have been heavily used for simulating such models in the field of computational neuroscience, yet AI-chips have not been tested on such workloads. The current paper aims at filling this important gap by evaluating multiple, cutting-edge AI-chips (Graphcore IPU, GroqChip, Nvidia GPU with Tensor Cores and Google TPU) on simulating a highly biologically detailed model of a brain region, the inferior olive (IO). This IO application stress-tests the different AI-platforms for highlighting architectural tradeoffs by varying its compute density, memory requirements and floating-point numerical accuracy. Our performance analysis reveals that the simulation problem maps extremely well onto the GPU and TPU architectures, which for networks of 125,000 cells leads to a 28x respectively 1,208x speedup over CPU runtimes. At this speed, the TPU sets a new record for largest real-time IO simulation. The GroqChip outperforms both platforms for small networks but, due to implementing some floating-point operations at reduced accuracy, is found not yet usable for brain simulation.

\end{abstract}


\section{Introduction} \label{sec:Introduction} To date, GPUs have achieved spectacularly better performance in deep learning (DL) than CPUs~\cite{reuther2019survey}. Recently, novel, specialized AI hardware platforms have begun to emerge, holding the promise of accelerating training and inference even further. The workloads targeted mainly are artificial, and specifically, deep neural networks (DNNs), which have shown great potential in recent years. On the other hand, highly biologically plausible models such as conductance-based (e.g., Hodgkin-Huxley) neurons have not attracted similar attention from AI-chip manufacturers and analysts alike. This is strange, given that biological brains -- the inspiration behind these DNNs -- are modeled using equations built on similar elementary functions. What is more, high-detail models are touted as the next AI wave, which is intended to be more biologically inspired than its predecessors~\cite{KASABOV2019111,MAASS19971659}. Therefore, it makes sense both for neuroscientists and for AI researchers to reach for these AI accelerators and deploy them for brain simulations; yet no performance studies exist.

In this work, we evaluate multiple, cutting-edge AI chips (Graphcore IPU~\cite{Graphcore_arch}, GroqChip~\cite{groq_arch}, TensorRT-capable GPU~\cite{nvidia_arch} and Google TPU v3~\cite{GoogleTUPv3_arch}) on simulating a highly biologically detailed model of a brain region, the Inferior-Olivary nucleus (IO). Biologically detailed brain models, such as the IO, chiefly involve addition, multiplication, division and exponential operations, arranged as sparse computations. There is, thus, a large operation overlap with artificial networks. Therefore, new AI chips seem a good fit for these types of models. This IO application represents timely, relevant research and is constructed as an extended-Hodgkin-Huxley model. It is very suitable for stress-testing the different AI platforms and highlighting architectural tradeoffs by adjusting the compute density, memory requirements and numerical accuracy of the IO model. Evaluation is performed using the application encoded as a TensorFlow~2~\cite{tensorflow2015-whitepaper} kernel, which in the case of the GroqChip, is necessarily compiled to its ONNX~\cite{bai2019} equivalent. ONNX is an intermediary tool used to convert models between different machine-learning (ML) frameworks.

\begin{figure}[t!]
    \centering
    \includegraphics[width=0.7\columnwidth]{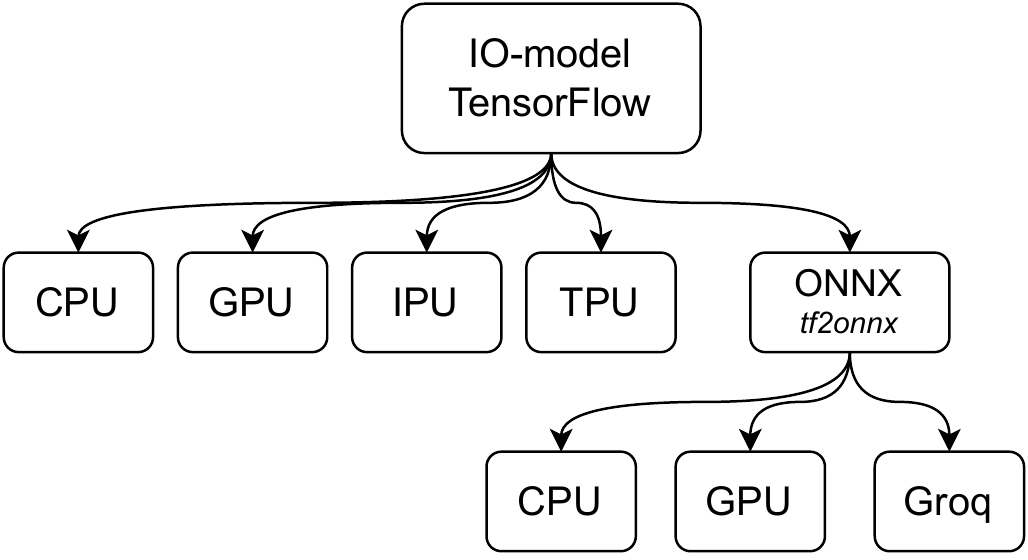}
    \caption{Overview of the performance-analysis strategy followed in this work}
    \label{fig:overview}
    \vspace{-0.4cm}
\end{figure}

In the analysis of the different accelerators, the exact same TensorFlow model is used, ensuring a fair comparison across the board. This is either used directly or ported to ONNX via the Python package \texttt{tf2onnx}. TensorFlow is a high-level API, requiring little to moderate intervention from the user and is therefore suitable for a wide user base. A schematic overview of this setup is presented in Fig.~\ref{fig:overview}.

While all accelerators in this study support chip-to-chip communication, this work constrains the application to single-chip performance comparisons; multi-chip is left as future work. The contributions of this work are:
\begin{itemize}
    \item We take a deep dive into four cutting-edge AI architectures with a focus on biologically plausible spiking neural networks (SNNs).
    \item We build the first ML-library-based, efficient implementation of a detailed brain model, the Inferior Olive (IO).
    \item We deploy the IO model onto the four AI platforms and benchmark their performance and numerical accuracy.
    \item We demonstrate that modern ML libraries are semantically able to model classical problems in scientific computing, offering large performance gains and reduced development times while remaining hardware-agnostic.
    \item Lastly, this work is the first to ever simulate a realistic mouse-sized IO model with real-time performance.
\end{itemize}

The paper is organized as follows: Section~\ref{sec:Related_work} presents related works in the field. Section~\ref{sec:IO_application} introduces the IO model used as our benchmarking application, while Section~\ref{sec:Target_platforms} briefly presents the four AI architectures under evaluation and attempts some performance predictions. Section~\ref{sec:Experimental_setup} ensures experiment reproducibility by detailing the experiment parameters and platform configurations used to acquire our results presented in Section~\ref{sec:Experimental_results}. A general discussion of our findings is included in Section~\ref{sec:Discussion} and conclusions in Section~\ref{sec:Conclusions}.

\section{Related works} \label{sec:Related_work} Models of biological neurons come in various levels of detail, ranging from population-level dynamics, from simplified models of single neurons to highly detailed biophysically realistic neurons~\cite{gerstner2012theory}. Coarse models of single neurons, notably leaky-integrate-and-fire (LIF) type models have seen a renewed interest in the DL-community (often referred here to as SNNs) as an alternative to artificial neural networks (ANNs)~\cite{tavanaei2019deep}.  In contrast, computational neuroscience is usually interested in biophysically accurate models that model the underlying biological processes in a way that makes it possible to gain insights about these processes. These conductance based models can be made more realistic by modeling of their 3-dimensional structure (the morphology) using multiple discretized compartments. Multi-compartmental conductance-based neurons are then simulated by explicit calculation of electrical currents flowing within, between and into discretized compartments~\cite{izhikevich2007dynamical}.

Due to the computional resources needed for large-scale conductance level brain simulations, computational neuroscience was an early adopter of general-purpose GPU (GP-GPU) in the HPC environment.
Notable GPU-based examples of large-scale, biologically detailed brain simulators include CoreNeuron~\cite{kumbhar2019coreneuron}, which enabled porting of existing conductance-level NEURON~\cite{carnevale2006neuron} models to the GPU, and more recently, Arbor~\cite{abi2019arbor}, a library-based approach to performance-portable, large-scale brain simulation. Their success shows that the computational problems of neuroscience map well to GP-GPU platforms and result in significant speedups for large-scale brain models. Still, even with hand-optimized CUDA code~\cite{vlag2019exploring}, the IO application (to be detailed in the next section) at biological sizes runs order-of-magnitude slower than the biological brain, hampering research.

With respect to TensorFlow-based implementations of conductance-level models, there is PymoNNto~\cite{vieth2021pymonnto}, an attempt to bring the Brian~\cite{goodman2009brian} API of neural models to TensorFlow. While faster than the Brian simulator on a GTX1080 GPU, performance was not a primary goal and the architecture prohibits optimizations using TensorFlow's JIT compiler backend, by scattering the computational definitions across the codebase.
Although this shows that TensorFlow \textit{does} express the right API surface for neural models, no efficient ML-library based conductance-level GP-GPU brain simulators exist.

Simplified SNN models
%, also of major interest to the DL community,
have readily available GP-GPU implementations of LIF and similar models as well. High-level ML-libraries like TensorFlow and PyTorch allowed for the hardware-agnostic implementation of their neural dynamics,
considerably lowering development efforts to build SNN simulators for GP-GPU simulation. For example, Nengo DL~\cite{rasmussen2019nengodl} allows for the GPU-based simulation of existing SNN models defined in the Nengo framework using TensorFlow.
Beyond just simulating neural networks on the GPU, novel developments in surrogate gradients for event-based SNNs and automatic gradient calculation provided by ML-libraries allowed for the nearly simultaneous appearance of similar SNN deep-learning libraries Norse~\cite{norse2021}, snnTorch~\cite{eshraghian2021training} and SpikingJelly~\cite{SpikingJelly}. BindsNET~\cite{bindsnet} is another, efficient SNN implementation in PyTorch with a focus on reinforcement learning.
Again, these project show that not only ML-libraries have the expressive power and performance needed to run large-scale SNN models, also that this arguably can be developed faster than hardware-specific low-level code. As these libraries had DL-oriented goals in mind, none of these implements multi-compartmental, conductance-level neural models. %thus not being useful for biologically realistic neuroscientific research.

Simplified SNN models also led to the development of specialized \emph{neuromorphic} hardware to simulate them.
%, promising energy and performance improvements.
Numerous publications show the benefits of using these chips for simplified-SNN simulation. For a short review of the various chips, we point the reader to~\cite{nawrocki2016mini}. While exciting with respect to low-power inference of SNN-based deep-learning models, these chips, due to their hardwired dynamics, lack the ability to simulate conductance level neural models.

On AI chips that have the semantic power to capture more general HPC workloads, little has been published about both simplified and conductance-level SNNs.
%The idea of using these DL-specialized chips for general-purpose compute is not new. For example, a single IPU achieves A100-comparable performance in a 2d Lattice-Boltzmann fluid simulation written in the Graphcore Poplar Framework~\cite{louw2021using}. This indicates that AI chips could run general HPC workloads, including brain simulations, at speeds at least comparable to GP-GPUs.
With respect to simplified SNN simulation, we find just one preprint targeting an AI chip, introducing an IPU-optimized version of snnTorch~\cite{sun2022intelligence}.
% These optimizations consist among others, of porting snnTorch's surrogate gradients to the automatic gradients calculation engine used on the IPU and training using the Graphcore-provided implementations of the Adam optimizer.
Training throughput of a dense 3-layer LIF network on an image classification task is 3.4x higher on the IPU than on the A100. The reported performance benefits decrease if the network size is increased, with the A100 apparently underutilized throughout the entire application. This shows the potential of using the IPU for simple SNN workloads, but the performance characteristics of other AI chips or more complex SNNs are not yet obvious.

No works have been published targeting AI chips with conductance-level models or other biologically realistic brain simulation scenarios, neither using high-level ML libraries or hardware-specific SDKs. To the authors' knowledge, this is the first work to implement an efficient, conductance-level, multi-compartmental neuron in an ML library and also the first to benchmark multiple AI chips on this workload class.

\section{The Inferior-Olive Application} \label{sec:IO_application} The IO is a
intrinsically oscillating
brain region located in the brainstem, and is key to motor control and learning~\cite{de2013inferior}. The estimated neuron population for the mouse brain is approx. $10^4$ neurons~\cite{yu2014inferior} and for humans between $10^6-10^7$ neurons~\cite{braitenberg2013cortex}. These numbers will be referred to during hardware-performance evaluation (Section~\ref{sec:Experimental_results}). In this work, we will capture in TensorFlow 2 the IO nucleus as an extended Hodgkin-Huxley (eHH) model, conductance-level brain model,
first published in~\cite{de2012climbing}. The model is a good example of the computational load of realistic brain models and, also, a good fit for our benchmarking purposes, since it captures complex neuron dynamics and fast interneural communication (in the form of gap junctions), as will be shown next.

We restate the IO-neuron main equations in this section,  but refer the reader to~\cite{de2012climbing} for more details. In addition, we model connectivity based on the network described in~\cite{negrello2019quasiperiodic}.

\subsubsection{The cable model}

\begin{align}
\begin{split}
        C_m \frac{dV^{(i)}}{dt} &=
       - \sum_{k \in \text{Channels}}I_k^{(i)}
       - \sum_{i \in \text{Compartments}}I_{k,j}
       \\
       &
       \phantom{.}
       - \sum_{i \in \text{Gap junctions}}I_{gj,k,j}
       - I_{\text{app}}^{(i)} \label{eq:voltint}
\end{split}
\end{align}

The eHH model describes the membrane that envelops the neurons as a capacitor. The cell internal voltage can thus be calculated by integrating currents
flowing into and out of the cell (eq.~\ref{eq:voltint}). Here, $I_{\text{app}}$ is an optional term describing externally applied currents
%by other connected brain networks or
by the experimenter.

\subsubsection{Channel currents}
Channels (CaL, h, KCa, Na, Kdr, K, CaH, Na, K) allow currents to flow through the cell membrane. They produce this current as function of internal state variables changing over time.
In general, this current (eq.~\ref{eq:chcur}) results from the potential difference to an channel specific reversal potential $E$ multiplied by the product of one or more internal gating variables, each optionally raised to an integer power (eq.~\ref{eq:chcur}).
The gating variables
%, limited $(0,1)$,
follow an Ordinary Differential Equation (ODE), that brings them to a certain cell-voltage dependent steady state at a given speed (eq.~\ref{eq:gating}). These latter equations are usually gaussian or sigmoidal functions of the voltage.
%Besides these general description,
For certain fast operating channels we set $n(t) = n^{\infty}(V)$ as a numerical stability optimization.

\vspace{-0.4cm}

\begin{align}
    I_i &= \bar g_i \left[\prod_k n_{i,k}(t)^{m_k} \right] (V - E_i)\label{eq:chcur} \\
    \tau_n\left(V\right)\frac{dn}{dt} &= n^{\infty}\left(V\right) - n\left(t\right) \label{eq:gating}
\end{align}

\subsubsection{Compartmental currents}
A single IO cell consist of three separate compartments, the axon, soma and dendrite. Currents flowing between different compartments are modeled resistively as:
$I_{i,j} = g_{i,j} \left(V_j - V_i\right)$

\subsubsection{Gap-junction currents}
Gap junctions are direct electrical connections between different IO cells and allow current to flow between them. They follow experimentally determined Connexin-36 protein dynamics:

\vspace{-0.2cm}
\begin{equation}
    I_{gj} = g_{gj} \Delta V \left[0.2 + 0.8\exp\left(-{\Delta V}^2/100\right)\right]
\end{equation}
with $\Delta V$ the potential-difference between two connected cells.

\subsubsection{Topology}
The real IO looks like a large, folded
%two-dimensional
sheet
%with both clustered and non-local neural connectivity
with mostly local %neural
connectivity.
As approximating this structure is not a focus of this paper,
%so to facilitate topology generation,
our model neurons are assumed to exist on a discrete 3-D grid with wrap-around connectivity (i.e., a hypertorus).
%Conversion of interneural distances in grid units to biologically meaningful units is left implicit.
%This is a good-enough representation of neural connectivity that
This should exhibit the same non-local memory-access patterns as a more realistic model.
% and as such not affect the benchmark.
Connections are sampled as a function of inter-neuron distance $r$ on a radially symmetric distribution:
$p(r) \propto u(r_{max} - r)(e^{-r^2} - e^{-r_{max}^2}) n(r)$,
where $n(r)$ is the density of neurons in the volume shell around $r$.
This distribution is sampled until we have 10 connections per neuron on average.

\subsubsection{TensorFlow Translation}
%\filbreak
% THESE NEED TO STAY TOGETHER OR THEY WIL BE UGLY
\begin{lstlisting}[style=EE,language=Python,label={lst:axna},
caption={Axonal sodium-channel current},captionpos=t,float,belowskip={-2ex},frame=top]
m_inf = 1/(1+tf.exp(-(V_axon+30)/5.5))
h_inf = 1/(1+tf.exp((V_axon+60)/5.8))
tau_h = 1.5*tf.exp(-(V_axon+40)/33)
dh_dt = (h_inf-h)/tau_h
I_na  = g_Na*(V_axon-V_Na)*m_inf**3*h
\end{lstlisting}

\begin{lstlisting}[style=EE,language=Python,label={lst:gj},
caption={Sparse gap-junction current},captionpos=t,float,belowskip={-2ex},aboveskip={0pt}]
Vdiff    = tf.gather(V_dend, gj_src) \
         - tf.gather(V_dend, gj_tgt)
I_per_gj = Vdiff * g_gj * (0.2 + \
             0.8 * tf.exp(-0.01*Vdiff*Vdiff))
I_gapp   = tf.tensor_scatter_nd_add(
             tf.zeros_like(V),
             tf.reshape(gj_tgt, (-1,1)), I_per_gj)
\end{lstlisting}

The previous equations sum up to a total of 14 ODEs per neuron. This system of ODEs is translated to a series of TensorFlow operators in Python. By defining the model in TensorFlow instead of using platform-specific APIs, we make sure all platforms have equal optimization opportunities. Furthermore, TensorFlow naturally translates to ONNX models, which is the only high-level API available for GroqChip.
Straightforward translation to TensorFlow is achieved by storing all state in a large 2d-array
and direct substitution of mathematical expressions by their TensorFlow counterparts (see Listing~\ref{lst:axna}).
% Naturally, we define brain simulation here as inference.
When certain model parameters need to be user-specified (e.g., $g_i$ or $I_{app}$), these are passed
%either as scalar or vector arguments
to the TensorFlow kernel,
which then needs to be recompiled before running again.

\begin{table*}[t!]
    \caption{Overview of all hardware used in experimental setups}
    \label{tab:machines}
    \begin{threeparttable}
        \begin{tabularx}{\textwidth}{lr@{~}lrrcc@{~~}l}
            \hline
            Device
                &  \multicolumn{2}{c}{On-chip Memory}
                &  Process node
                &  Transistor count (Bn)
                &  Base-boost freq. (MHz)
                &  TDP (W)
                &  Software
                \\
                \hline
            AMD 3955WX CPU *
                &  128 GB & DDR4
                &  7 nm
                &  19.94
                &  3900 - 4300
                &  280
                &  TF 2.11.0
                \\
            GroqChip TSP
                &  230 MB & on-chip
                &  14 nm
                &  26.8
                &  900
                &  -
                & Groq SDK 0.9.1 ***
                \\
            % Nvidia RTX 6000 GPU
            %     &  24 GB & GDDR6
            %     &  12 nm
            %     &  18.6
            %     &  1440 - 1770
            %     &  260
            %     &  ONNX Runtime 1.13.0 ***
            %     \\
            Nvidia A100 GPU
                &  80 GB & HBM2e
                &  7 nm
                &  54.2
                &  1275 - 1410
                &  400
                &  TF 2.11.0
                \\
            Graphcore IPU (GC200) **
                &  900 MB & on-chip
                &  7 nm
                &  59.4
                &  1330
                &  185
                &  TF IPU 2.6.3+gc3.0.0
                \\
            Google TPUv3
                &  32 GiB & HBM
                &  16 nm
                &  (est.) 11
                &  940
                &  450
                &  TF 2.11.0
                \\
                \hline
        \end{tabularx}
        \vspace{0.1cm}
        \textit{*AMD Ryzen Threadripper PRO 3955WX (16-Core) $|$
        **Single M2000 in IPU-POD16 (with 4 GC200 chips) $|$
        ***TF2ONNX 1.13.0 and ONNX opset 16}
    \end{threeparttable}
    \vspace{-0.3cm}
\end{table*}

Translating gap junctions to both TensorFlow and ONNX in a performant way requires expressing them as vector operations, as opposed to more traditional \texttt{for-loop}-based approaches~\cite{vlag2019exploring}. With just 10 connections per IO neuron on average, cell-to-cell communication is sparse. The effective operation from a TensorFlow perspective is two sparse-matrix (SM) multiplications. As a novel contribution in computational neuroscience, we model those as \texttt{tf.gather} and \texttt{tf.tensor\_scatter\_nd\_add} operations (see Listing~\ref{lst:gj}). Apart from being more specific and memory-efficient in describing SM multiplications, these functions have a direct mapping to ONNX operators as \emph{Gather} and \emph{ScatterND} since ONNX specification \texttt{opset 11}, contrary to SM multiplications which currently are not possible in ONNX.

At each timestep, ODEs are integrated using Forward-Euler to produce the next state array, resulting in a hardware-agnostic timestepping function. For TensorFlow backends, a JIT-compilable TensorFlow function is constructed that executes 40 timesteps at a $\Delta t$ of $0.025ms$, resulting in a $1ms$ sampling accuracy. For ONNX backends, the timestep function is converted to an ONNX model and either the public onnx-runtime library or Groq compiler is used to compile this into executable code. This does not lead to the best possible performance by default, thus hardware-specific optimizations are discussed in Section~\ref{sec:hwopt}.

\section{Target Platforms} \label{sec:Target_platforms} Hardware platforms were selected from the top-performing AI accelerators in the MLCommons MLPerf training benchmark v2.0~\cite{mlcommons}. From this, the Intel Habana Gaudi was not available to us. The GroqChip was included as it was already available through academic channels. An overview of all AI chips is given in Tab.~\ref{tab:machines} and will be presented next. A modern, server-grade CPU is also included as a baseline for our subsequent performance and numerical-accuracy comparisons.

\subsubsection{\textbf{Nvidia GPU}~\cite{nvidia_arch}}
These are well-established accelerators in the HPC world. With the introduction of the Tensor Cores in Nvidia GPUs, they also became well-known for their AI capabilities. Tensor Cores are capable of matrix multiplications in a very efficient manner. The current generation of tensor cores can support up to TensorFloat-32 (TF32) precision TF32 is a floating point with float 32 dynamic range but float 16 precision. There are multiple ways of interacting with them; e.g., via cuBLAS and TensorRT.

\subsubsection{\textbf{GroqChip}~\cite{groq_arch}}
This is a deterministic Tensor Streaming Processor (TSP), resembling a modified systolic array architecture. The chip layout is a conventional 2D mesh of cores, each with its own dedicated functionality. A column of these cores -- all of the same type -- is called a functional slice. Data travels horizontally, executing 320 SIMD-style lanes. A single instruction can control 16 lanes, effectively creating 20 superlanes that can all be operated independently from each other. The functional slices consist of one vector processor (VXM), two matrix execution modules (MXM), switch execution modules (SXM) and memory modules (MEM). Each functional unit (core) accepts a set of instructions; for example, the MEM unit could receive the instruction to put a vector onto one of the data streams or store the results from the data stream in its available SRAM. As soon as data is loaded onto a data stream, it automatically `flows' in the direction of the stream, which can be either EAST-bound or WEST-bound. When an addition needs to be performed, both inputs need to arrive at the same time as the \texttt{add} instruction at the corresponding VXM core. This design choice puts the burden of optimization on the software generating the instructions. This is either done by the Groq compiler automatically from an ONNX-graph input or manually controlled by a user through the exposed Groq-API, which has different levels of abstraction on top of the Groq-ISA. To support the creation of large-scale systems, the GroqChip has dedicated Chip-to-Chip modules that are capable of performing off-chip communication without losing their determinism~\cite{groq2022}. For this work, we will mainly utilize the VXM and MEM units, The memory modules add up to a total of 220 MiB of on-chip SRAM. Each superlane implements a 4x4 mesh of vector ALUs capable of doing x16-SIMD. Each ALU has a 32-bit input operand but with the exception of additions and multiplications, instructions are done in a reduced-precision FP32 format.

\subsubsection{\textbf{Graphcore IPU}~\cite{Graphcore_arch}}
The Graphcore Intelligence Processing Unit (IPU) is designed for efficient execution of fine-grained operations across a large number of parallel threads. By design, the IPU offers true Multiple Instruction, Multiple Data (MIMD) parallelism. This unique style of parallel-processor design adapts well to fine-grained computations that exhibit irregular data-access patterns. Each IPU contains 1,472 tiles, containing 1 core and 624KiB of SRAM memory. A single core can only access the memory in its own tile. Intra-IPU communication relies on a powerful, low-latency interconnect called \texttt{IPU exchange}. For inter-IPU communications, each chip contains 10 so-called \texttt{IPU links}. The IPU compute units, called \texttt{Accumulating Matrix Product} (AMP) units, support FP32 arithmetic and are designed to accelerate matrix multiplications and convolutions. With respect to the programming model, the IPU adopts the Bulk Synchronous Parallel (BSP) model~\cite{BSPmodel} through which it organizes its compute and data-exchange operations. This abstraction for parallel computations consists of multiple sequential supersteps. A superstep consists of a local computation phase; every process (tile, in the IPU case) operates in isolation performing compute only on its local memory, followed by a communication phase where each process can exchange values needed by other tiles. These activities are concluded with a barrier synchronization phase; only when all processes have reached the barrier can the next superstep be started. Because of this, the IPU can be described as a true BSP machine.

\subsubsection{\textbf{Google TPU}~\cite{GoogleTUPv3_arch}}
The TPU (version 1) was designed as a systolic-array processor for inference, only supporting 8/16-bit operations. By supporting only matrix-multiply and basic nonlinear activation functions, it was unfit for training neural networks. Consequentially, an HPC application -- for example, the one demonstrated in this paper -- would also not be a suitable fit for this processor. However, with the TPUv2, Google shifted their focus towards supporting training on their TPU chips. Google added a vector-processing unit (VPU) and changed the matrix-multiply units to support the FP16 format (FP32, with only a 7-bit mantissa). The VPU most likely supports higher precision, as can be deducted from results in this work but no confirmation of this is found in the public domain. These two major (micro)architectural changes made it possible to run a wider range of applications including training neural models on the TPU. All are supported through the Google XLA compiler taking TensorFlow as input. The TPUv3, assessed in this work, is an upgrade in terms of functional-unit count, higher memory speed, and optimized chip layout, but did not include any fundamental changes.

\subsection{Performance Predictions} The IO application has two components that map differently onto different types of hardware: i) a part with embarrassingly parallel computations for updating local neuron states; and ii) a part with SM computations for exchanging membrane voltages over gap junctions. Before we proceed to the actual experiments, we attempt performance predictions, driven by the idiosyncrasies of the different AI-chip architectures.

\subsubsection*{Embarrassing parallelism}
These are calculations for updating the state of every single neuron. This boils down to elementwise vector operations. The GPU architecture featuring one Warp execution per Streaming Multiprocessor or multiple Tensor Cores is very well-suited for this type of parallelization. The TPU and the GroqChip are both based upon systolic-array architectures, both natively supporting Matrix-Multiplication but also Vector-Operation operations that can be utilized for these calculations. In fact, since neuron updates require only ~1-D~ data, the Matrix-Multiplication units (which is the focus of these chips) are effectively %being
underutilized in these architectures. The IPU, with a large amount of very small general-purpose cores, should also do well on parallelizing neuron-state calculations, however, its architecture is geared towards irregular data-access patterns, which is not essential to the particular task. The extra overhead of such advanced features, therefore, will not help performance in terms of computing this embarrassingly parallel part of the simulation.

\subsubsection*{Communication}
As described previously, gap-junction communication employs the gather-scatter operations (essentially, SM operations) from TensorFlow. For either the GPU or the IPU, such operations are handled better due to the different execution paths that can be handled within the architecture by design. In contrast, the GroqChip and TPU need to handle these differently: a naive approach would be to enforce dense-matrix operations via one-hot encoding of operands and, then, utilizing the matrix-multiplication hardware. In case the GroqChip or the TPU happen to use such a strategy, we expect that performance will deteriorate very rapidly or memory will be depleted with increasing IO-network sizes.

\subsubsection{CPU + TensorFlow}
For this platform, JIT compilation through the XLA compiler~\cite{xla} will be used; it will automatically utilize the many threads nowadays available in CPUs. We expect decent performance and very accurate results because of full FP32 support. Since it is the hardware on which brain models are traditionally executed and gives accurate results, the CPU will form our baseline. Accelerators should outperform this implementation in terms of runtime, especially for larger network sizes.

\subsubsection{GPU + TensorFlow}
The XLA compiler is used, which optimizes the graph resulting in a single kernel launch. Among others, it does this by ``fusing'' the calculations. Moreover, this fusion keeps intermediate values stored in GPU registers~\cite{xlabloggpu}.
The TensorFlow backend for CUDA use Tensor Cores, at a loss of FP32 accuracy.
However, this only happens when explicit matrix-multiplications are requested and not as an optimization. So in our case, the compiler will only use float32 CUDA operations.

\subsubsection{IPU + TensorFlow}
The IPU architecture is not a perfect fit for the embarrassingly parallel part of the computation. For the interneuron-communication part, the BSP model is a better fit and thus is expected to perform better.
However, as the topology is given as an unknown parameter to the model, the IPU compiler can not be expected to allocate neighboring cells on adjacent tiles, resulting in sub-par communication performance. Available memory should easily be able to handle large problem sizes.

\subsubsection{TPUv3 + TensorFlow}
The TPU supports FP32
%and is a systolic-array processor
and is expected to handle our workload, especially for the unconnected case, very well. As Google put much effort into TensorFlow support, gather-scatter operations are expected to be optimized, to the best of the hardware capabilities. Because of FP32 support in the v3 model, we expect correct numerics in the output, as well.

\subsubsection{CPU/GPU + ONNX}
Expectations are the same as for CPU/GPU + TensorFlow. We expect the XLA compiler to outperform the ONNX runtime slightly for the CPU case simply because it can perform whole-program optimization.
For the GPU, this effect is expected to be much larger and the TensorFlow is expected to dominate ONNX as the ability to fuse kernels will be a big advantage for TensorFlow over single-kernel invocations in ONNX. Especially the invocation overhead for small GPU kernels will hurt the performance of the ONNX-GPU-runtime. TensorRT is also a supported backend in ONNX that is expected to outperform the CUDA runtime in performance; it will, however, drop precision as the backend switches to TF32 numerics.

\subsubsection{GroqChip + ONNX}
The GroqChip is a new, upcoming modified systolic-array processor. Its compiler takes in the ONNX graph but is not limited to executing this on an operation-per-operation basis as it recompiles the full ONNX graph at once. Therefore, it can potentially perform the same optimizations as the XLA compiler for the TPU. 
As the first version of the architecture, current compiler development is still exploring ways to map non-standard ML-operations to the hardware.
Besides, the GroqChip VXM is not capable of doing all operations in IEEE FP32 arithmetic. Because of this, it can be expected to perform slightly better than the TPU at the cost of reduced accuracy. 

\section{Experimental Setup} \label{sec:Experimental_setup}
\subsection{Benchmarking Parameters} Each platform is benchmarked for \emph{performance} on a set problem (i.e., network) size as well as for its \emph{performance scalability} by simulating the IO network for small population sizes in the range $[4^3, 5^3, \ldots, 20^3]$ and, again, for larger sizes in the range [$30^3, 40^3, \ldots, 100^3$], where the third power is an artifact of the cubic network-topology generation method. These experiments are focusing on four different aspects of each AI platform, discussed next.

\subsubsection{Unconnected Network}
By removing the communication step (gap junctions) from the model, we obtain a (biologically unrealistic) compute-heavy, embarrassingly parallel workload. First, we measure the setup time for each AI platform, including on-chip buffer allocation, Ahead-Of-Time (AOT) compilation or definition of Just-In-Time (JIT)-enabled functions. Next, we simulate an IO network for 100ms of biological time and take the minimum wall-clock time from 5 runs (including data-transfer times). For JIT targets, the first runtime (if outside the other runtimes' standard deviation) minus follow-up runtimes is taken as the JIT compilation time, such that we can compare setup times between AOT and JIT targets.

\subsubsection{Connected Network}
By restoring gap junctions into the IO network, we assess communication overhead. Runtimes are obtained in an identical way as before, yet the expectation here is that they are markedly longer than the unconnected case.

\subsubsection{Numerical Validation}
Measuring performance is our main focus, yet this must not come at the cost of functional correctness. Here, we simulate connected networks up to 729 neurons for 10 seconds of biological time and numerically compare the various results to the reference CPU output.

\subsubsection{Numerical Stress-test}
Here, we simulate the IO in a more biologically realistic way that is of interest to neuroscientists: We add more variance to the neural parameters and, most importantly, a lot of external current inputs (simulating other brain regions) that will evoke action potentials (spikes) in the IO dynamics. These fast transients will stress-test the numerical performance of the AI hardware, especially non-IEEE754 targets (Tensor Cores and GroqChip). We perform this experiment on the smallest 64-neuron network and then compare for numerical accuracy against the CPU.

Benchmarking is implemented in a publicly available and modular, extensible framework, downloadable from GitLab
\url{https://gitlab.com/neurocomputing-lab/Inferior_OliveEMC/ioperf}.
The main benchmarking script auto-discovers available hardware, runs the appropriate benchmarks and records results. Used software versions are also shown in Tab.~\ref{tab:machines}. 
\subsection{Hardware-specific Optimizations} \label{sec:hwopt} \begin{figure*}[t!]
    \centering
    \includegraphics[width=0.9\textwidth]{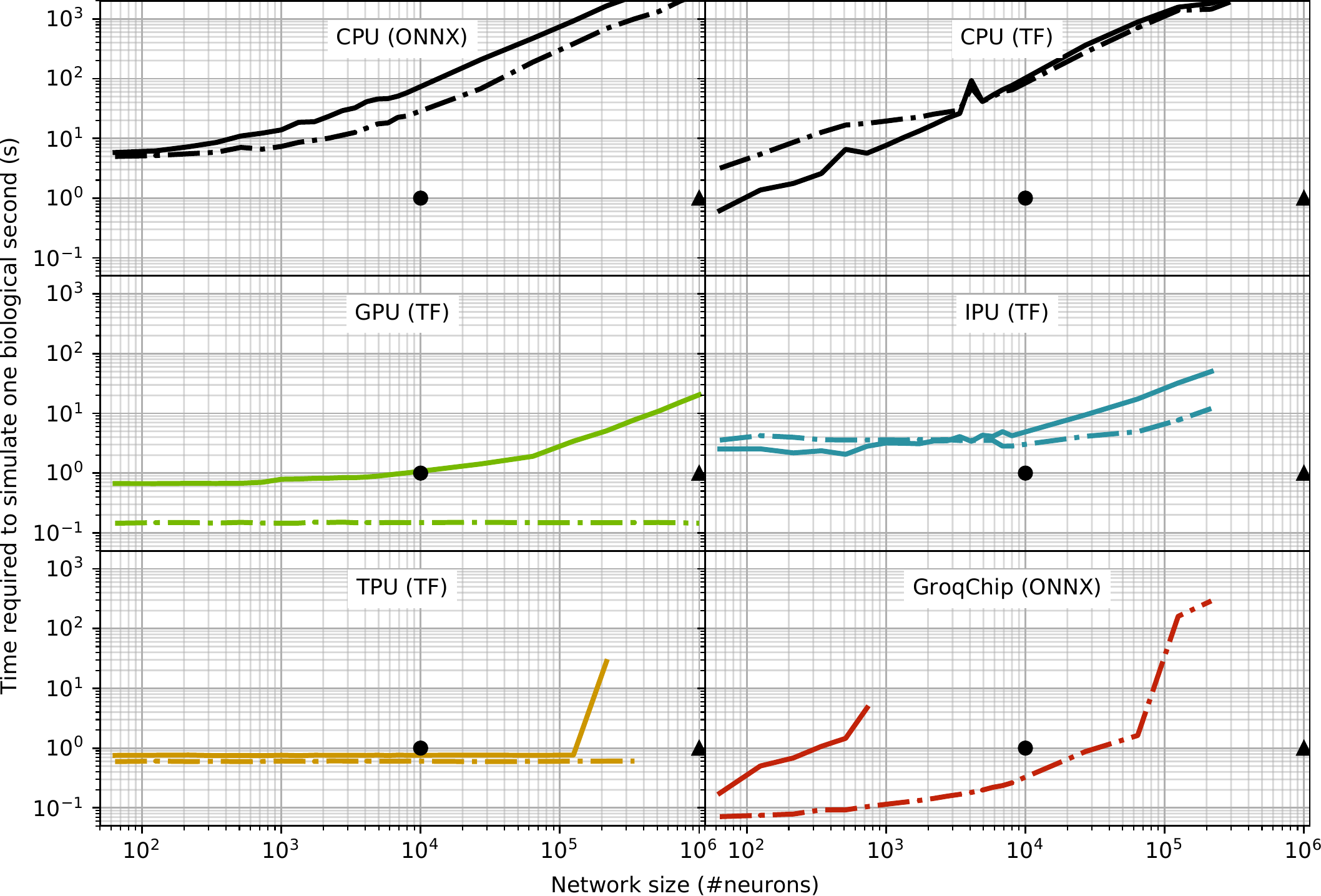}
    \caption{Runtime performance (lower is better), comparison between CPU baseline, GPU and AI chips. For scale, the mouse ($\protect\bullet$) and human ($\protect\blacktriangle$) Inferior Olive are shown as running in realtime in all figures. The CPU is included twice to explain the observed switching behavior of the IPU. On the CPU, while the XLA optimizer builds a single-core, connected-network simulation, it builds a multicore, unconnected-network simulation (as observed by load-testing), leading to an unexpectedly \emph{slow} simulation for the latter case. The same behavior can be observed for the IPU, which uses the XLA compiler as well. 
    }
    \label{fig:mainfig}
    \vspace{-0.4cm}
\end{figure*}

While our original goal was not to write platform-specific code, we found that by default some of the AI platforms did not perform very well. % on the I/O.
For example, most platforms defaulted to copying over the entire parameters arrays for each kernel invocation, which was not needed for this mostly constant data.
For a fair comparison between hardware platforms, we allowed optimizations to be applied to hardware-specific code that either led to operation fusion across different execution kernels or prevented unnecessary device-host data transfers. The exact optimizations have been applied in close collaboration with Graphcore and Groq for the respective chips, and are as follows:

\subsubsection{TensorFlow XLA}
The TensorFlow graph executor typically performs each operation separately when a graph is run with a corresponding kernel invocation. A different way to run TensorFlow models is made available by XLA, which turns a TensorFlow graph into a series of kernels created for a particular application. These kernels can take advantage of application-specific information for performing optimizations, e.g., operation fusion. The CPU, GPU, and TPU are the three available backends for the XLA compiler. For the IO application, a TensorFlow wrapper function was implemented that fuses up to 40 timesteps together for each call in order to fully exploit the XLA compiler.

\subsubsection{ONNX}
Except for the GroqChip, all ONNX implementations build on top of \emph{onnxruntime} or \emph{onnxruntime-gpu}. We enable all backend-supported graph-optimizations. Explicit \texttt{IOBindings} are used to prevent unneeded host-device data copies. Parameters are copied once to the device at simulation start. Then, state is allocated twice, with each timestep toggling between two buffers, one as the input state and the other as the output (next) state. For TensorRT, we leave the default behavior of using TF32 enabled, otherwise, it will not utilize its Tensor Cores.

\subsubsection{Groq}
After the compilation of an ONNX graph with the Groq Compiler, the binary can be executed directly on the GroqChip. A naive approach here would be to invoke this binary 40 times for 40 timesteps and move the data back and forth continuously since the GroqChip only has SRAM which is fully managed at compile time. However, the Groq Compiler is able to tie input and output tensors together into a \emph{persistent} memory buffer in the on-chip SRAM. Utilizing this results still in 40 invocations of the binary but skips the continuous I/O between host and accelerator. A more radical way to improve the performance is to compile the 40 timesteps into a single ONNX graph that can then be converted with the Groq Compiler; this method will reduce 40 invocations to a single invocation. We implemented all optimizations as long as the compiler was able to compile them. The 40 timesteps at once quickly ran into compiler errors with growing networks.

\subsubsection{Graphcore}
The IPU has architectural support for \emph{streaming memory}. This means that we can run a single program on-chip for the entire simulation that will stream out samples every 40 timesteps. The inner, unsampled 1-msec 40-timestep loop, is run using \texttt{ipu.loops.repeat}, after which the recorded voltages are pushed to an \texttt{IPUOutfeedQueue} with a 200-sample size. This is, then, looped once more using \texttt{ipu.loops.repeat} for the required amount of milliseconds to simulate and wrapped in a TensorFlow JIT function. Furthermore, the fast-math optimization is enabled, 128 IPU tiles are reserved for I/O with \texttt{place\_ops\_on\_io\_tiles = True} and program execution is limited to a single IPU.

\section{Experimental results} \label{sec:Experimental_results} With the exception of the reference CPU, for brevity we report here either TensorFlow or ONNX results, depending on which of the two leads to better performance. Overall performance plots are shown in Fig.~\ref{fig:mainfig} and will be detailed in the next sections. In general, it is found that, for the IO application, the ONNX ports are outperformed by their TensorFlow counterparts. This is due to the fact that the onnx-runtime library currently does not perform as extensive optimizations as the XLA compiler.
For example, the CUDA target translates each compute step into a single predefined kernel call. The TensorRT backend performs operator fusion, resulting in multiple kernels that chain arithmetic operations. Still, the CUDA XLA-backend vastly outperforms both ONNX CUDA targets, and as such we removed the corresponding findings from the main analysis. Note that the Groq platform only supports AOT compilation of ONNX models.

\begin{figure}[t!]
    \centering
    \includegraphics[width=\columnwidth]{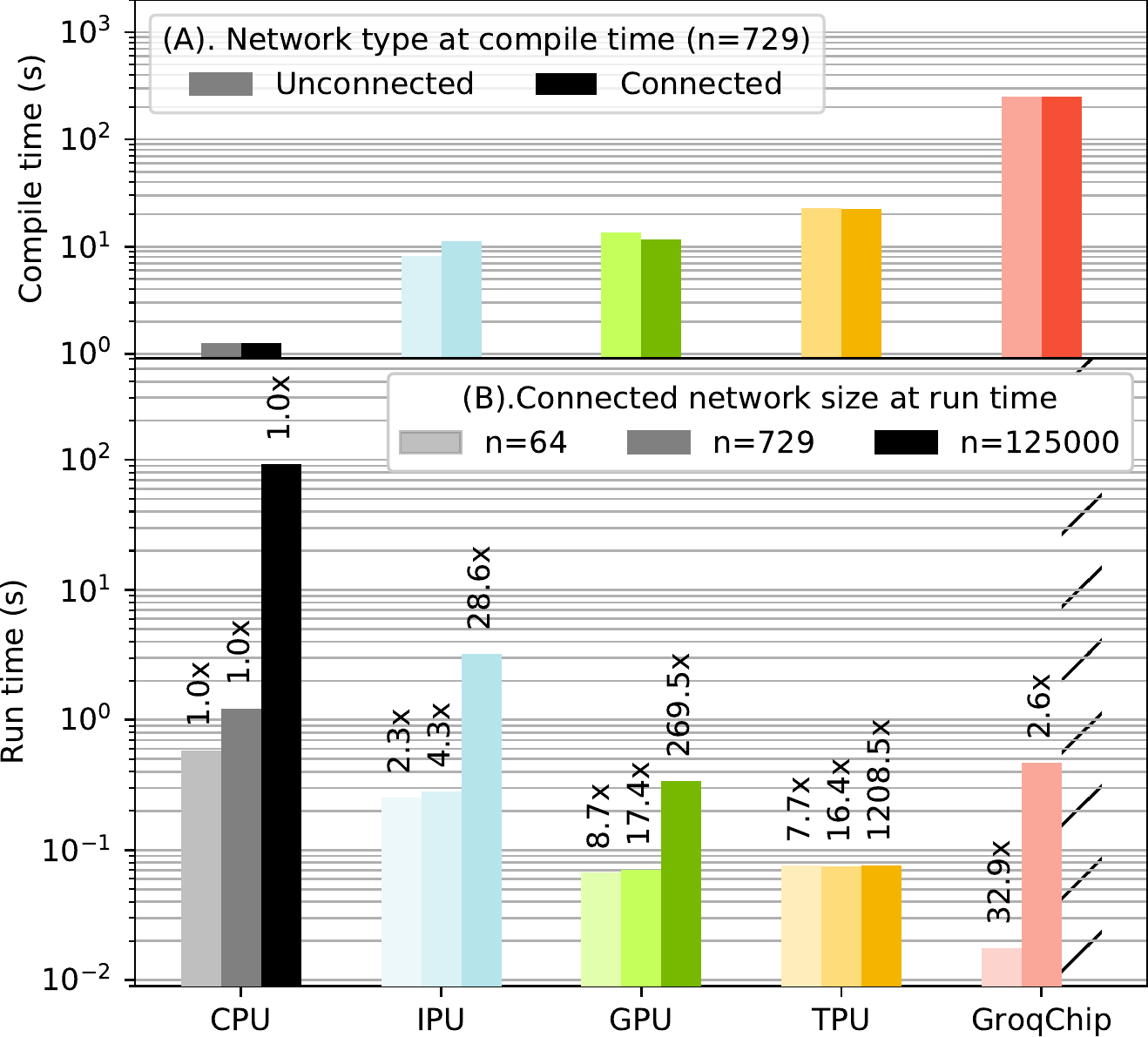}
    \caption{\textbf{(A)} Setup (AOT+JIT compile + memory allocation) times for a network of 729 neurons in both Unconnected- and Connected-network configurations. JIT compile times are extracted from the first run of 5 performance runs and added to the initial setup time (if outside one standard deviation). \textbf{(B)} Performance and speedup of different AI chips vs. the CPU reference on the Connected benchmark, for different network sizes. Sizes were chosen to be the smallest (64) and largest connected networks that could fit on the GroqChip (729) and the TPUv3 (125,000). The rightmost GroqChip bar is absent, corresponding to the model that could not be compiled.}
    \label{fig:ctime}
    \vspace{-0.4cm}
\end{figure}

\subsection{Compilation Time}
Both software stack and hardware influence program setup time, as illustrated in Fig.~\ref{fig:ctime} for the largest network (729 cells) that could fit in all AI chips. The CPU compiles the fastest across the board as we have a direct translation of ONNX operations to their CPU-optimized callbacks. The TensorFlow (XLA) version, not included in the figure, was much slower due to the increased compiler complexity. Both the IPU and GPU exhibit similar JIT compilation speeds. The GroqChip's AOT compiler takes significantly longer for this workload due to the explicitly concatenated 40 timesteps. The GroqChip version with a single timestep per program compiles much faster than the Graphcore or A100 versions, but at a small performance loss.

\subsection{Runtime Performance}
\subsubsection{Unconnected Network (embarrassingly parallel)}
For unconnected cells, neural dynamics are expressed only using vectorized operations. As predicted, this fits the compute paradigm of the GPU very well. Performance scales linearly with problem size (horizontal line), showing that the GPU cores are underutilized for all simulated network sizes.

The TPU and GroqChip, as systolic-array-based processors, were expected to be a poorer architectural fit because large parts of the chip would be left unused. Still, the focus on efficient vector operations could result in speedups. We can indeed observe this in Fig.~\ref{fig:mainfig}, although in different ways. The TPU, similar to the GPU, flatlines across all problem sizes,
although being 4.1x slower. Consequently, memory capacity is not a problem for the TPU but performance capping in raw single-cell computations due to architectural design choices. In contrast, the GroqChip starts out 2.0x faster than the GPU, quickly loses this edge and, between $10^3$ and $10^4$ cells, starts to hit its memory-capacity limits, degrading performance with higher problem sizes. Networks of more than $640,000$ cells simply do not fit on the chip. The GroqView analyzer confirms that the problem is core-to-core-memory communication and that most dedicated cores are not used most of the time.

The IPU was expected to perform well given its large core count but the very homogeneous compute load proved a poor fit for its MIMD design, leading to large under-utilization of the chip. With respect to real-time performance, only the GPU followed by the GroqChip (ignoring memory issues) and marginally the TPU makes the 1-sec cut.

\subsubsection{Connected Network (high communication overhead)}
As predicted, communication patterns induced by a small number of gap junctions lead to a large performance reduction of 4.6x for small networks on the GPU. For higher problem sizes, performance drops at a growing rate, with a 141x degradation for networks of $10^6$ cells. The AI chips fare much better here, most of which initially shows a less than 20\% reduction in performance against their unconnected counterparts.

As an exception, the GroqChip's connected-network simulation runs 2.5x slower than the unconnected version; even so, it outperforms the GPU on very small, connected neural networks by a 3.7x speedup. However, the GroqChip (as expected) converts the SM communication into a dense-matrix multiplication, making the best out of its deterministic-execution hardware. This quickly leads to prohibitively large matrix multiplications and, beyond 729 cells, the scheduler is unable to allocate the necessary instructions. In effect, the GroqChip loses its edge over the GPU for larger networks.

In contrast, the TPU shows nearly identical behavior to the unconnected case and its performance does still not scale with problem size. This changes around networks larger than $10^5$, where the \emph{JIT compiler} seems to run into performance problems. Here, we observed large random fluctuations in performance that either led to approx.\ 1-sec or very long more than 400-sec run-times over the 5 repeated runs.
We expect that these originate from memory limits of the TPU and had to stop benchmarking due to impractically large run times. However, we could not determine the true source of variation.

The IPU, severely underutilized for the unconnected case, sees in fact a performance improvement when we increase the communication overhead in small networks. While counterintuitive, this is actually the same effect we see on the XLA-based CPU backend. Here, we see that gap junctions force the simulation to become single core, which actually becomes faster than the parallel, multi-core, unconnected case, due to the lower synchronization overhead. Around $10^4$ cells, this behavior changes, gap-junction communication becomes a fixed overhead on top of normal simulation. At a certain point, this growth becomes exponential and the largest simulated network does not fit on a single IPU anymore.

\subsection{Numerical Validation}

\begin{figure}[t!]
    \centering
    \includegraphics[width=\columnwidth]{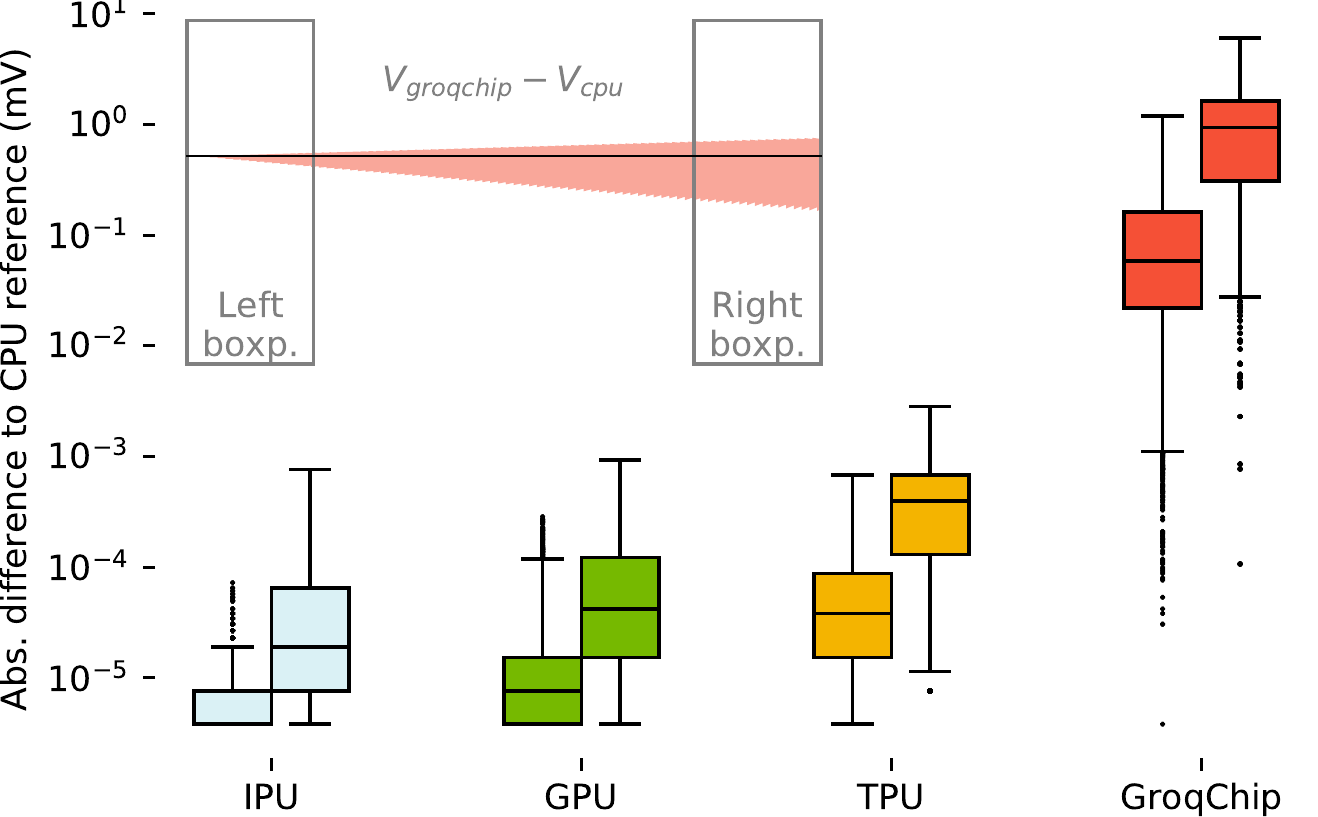}
    \caption{Numerical-accuracy validation (lower is better). Box plots show deviations from CPU baseline, as recorded over two 1-sec timespans, one at the start (left) and one at the end (right) of the 10-sec numerical-validation simulation. The GroqChip result, showing the largest deviation, is plotted in the upper left corner together with the two recording spans .
    }
    \label{fig:valacc}
    \vspace{-0.5cm}
\end{figure}

While all AI chips outperform the CPU baseline, it is wise to explore also any potential decrease in numerical accuracy of the different runs with respect to that of the same CPU. Here, we compare 1-msec sampled cell somatic voltages of an extended, 10-sec simulation for a 729-cell, connected network (the largest population supported by all platforms); results are shown in the box plots of Fig.~\ref{fig:valacc}.

As expected, platforms supporting IEEE754 floating-point numerics (IPU, GPU, TPU) show accurate reproduction of voltage traces. The IPU, even with fast-math enabled, is the most faithful to the CPU baseline. The GPU and TPU exhibit increasingly large deviations but still fall within limits explainable by floating-point instruction reordering. 
The GroqChip, while supporting FP32 number storage, implements certain operations at lower precision including exponent calculation. This is visible by a quite large $mV$-order deviation from the CPU baseline, for a process that happens at the $10-100mV$-scales. This voltage difference mostly stems from a slowly accrued phase difference for the oscillating cells. TensorRT (not shown in this plot) is by default using Nvidia's TF32, for which accuracy was found similar to that of the GroqChip.

\subsection{Numerical Stress-test}

The numerical stress test increases neuronal variation and adds external inputs that lead the neurons to spike.
These fast transients can not be simulated using FP16 precision, but reduced-accuracy FP32 operations as used in Tensor Cores or GroqChip are still untested.
Once more, we compare the deviation of the somatic-voltage traces of the various AI chips against the CPU baseline.

Again, the platforms with native FP32 support show the lowest deviation: For the IPU this is $0.087 mV$, for the GPU this is $0.135 mV$ and for the TPU this is a $0.672 mV$ maximum absolute difference from the CPU baseline. These moderate, $mV$-order differences can be explained by small spike-time differences which due to the large neuronal-spike sizes quickly lead to large voltage discrepancies. Importantly, all simulations run stably; i.e., do not cause this chaotic IO-model simulator to crash. The GroqChip simulation initially starts out the same as in the numerical-validation test, but as soon as input perturbations are applied, it becomes unstable and settles on voltage deviation at a measured maximum of $8.51\times 10^{36} mV$, unacceptable for scientific applications. Notably, the error stabilizes at this point and does not explode to infinity or $NaN$ values, as observed with FP16 simulations. To regain numerical stability, we tried lowering the time-stepping constant $\Delta t$ 10-fold and 100-fold for the GroqChip simulation, but this did not lead to results more closely in range with the CPU ones.

\section{Discussion} \label{sec:Discussion}
As this work has shown, utilizing AI platforms for executing highly biologically plausible SNN workloads is made exceedingly user-friendly when using a ML-library like TensorFlow. Arguably, even better performances could be obtained by coding via the various hardware SDKs (Software Development Kits), but it is unrealistic to expect computational scientists to learn the low-level details of all hardware options made available to them these days.

As shown, the added benefits from JIT compilation make a hand-coded CUDA implementation perform on par with the XLA-compiled TensorFlow version while, at the same time, allowing one to move easily to a new piece of hardware when this is released. We expect that, in the future, more classical HPC workloads will see ML-library, that is, tensor-based implementations.

For promising upcoming accelerators like those by Graphcore and Groq, we believe that future speedups will chiefly come from software and compiler upgrades, as current SDKs are mostly optimized for ML workloads. For instance, gather-scatter operations on the GroqChip do not have to be implemented as dense-matrix operations, memory can be better utilized, and better support for iterative programs must also be introduced. The TPU which is architecturally similar to the GroqChip, clearly performs gather-scatter operations in a more efficient way than encoding indexing as one-hot vectors.

Speed-ups could be gained by effective use of mixed precision on the IPU or reduced accuracy FP32 operations using Tensor Cores or GroqChip. For the IPU, this would constitute a separate numerical sensitivity analysis to find out which parts of the compute graph can be lowered to (stochastic rounded) FP16. As shown, the accuracy loss on Tensor Cores and GroqChip does in its current form not allow for brain simulation, but possible these could be put to use by switching the integration scheme or other numerical optimizations.

Finally, this work has steered clear off multi-chip topologies. All discussed architectures do support specifically developed, low-latency, chip-to-chip hardware and assorted communication protocols. In many ways, such coherent communication is a bigger and more timely challenge than acceleration speed itself, which would deliver massive benefits for large-scale SNN simulation (or training). However, tapping into those platform-specific interfaces requires SDK-specific coding of the IO application; relying on TensorFlow or ONNX frameworks will, generally, not work. Careful and platform-specific coding is necessary, which we leave as future work.

\section{Conclusion} \label{sec:Conclusions}
In this work, we built the first ML-library-based, efficient implementation of a large-scale, conductance-level brain model, the Inferior Olive (IO). Subsequently, we benchmarked the performance of simulating this model on a 16-core AMD Ryzen Threadripper PRO 3955WX CPU, an Nvidia A100 GPU, and different AI chips (Graphcore IPU M2000, GroqChip and Google TPU v3). We found that all accelerators provide significant speedups over the CPU implementation. For this specific problem, the GPU and TPU seem most fit for simulation, with the TPU setting a new record for real-time IO simulation. For small networks, the GroqChip outperforms the other accelerators, but large networks could not fit in the on-chip instruction memory. More generally, we hypothesize that modern ML-libraries possess the semantic power to model classical problems in scientific computing. These, then, map extremely well to ML-driven, novel AI-chip architectures, which apart from large performance benefits, also benefit from reduced development times. For example, the version of our IO application running on the TPU outperforms the handwritten and hand-optimized CUDA implementation by a large factor, at a fraction of the development cost. The exact hardware trade-off will vary on an application-by-application basis, and hardware selection also benefits significantly from the hardware-agnostic model description.

\section*{Acknowledgement}
This research would not have been possible without
access to dedicated hardware:
The RTX6000 was gifted from the NVIDIA Hardware Grant Program,
Google provided free cloud credits for TPU access
and Graphcore provided access to POD16 machines through
Paperspace and Gcore cloud via its Academia program.
Furthermore, we'd like to thank Graphcore
employees for helping with optimizing the IPU code
and Dr.\ Mario Negrello for neuroscientific insights.

\bibliography{main}{}
\bibliographystyle{IEEEtran}

\end{document}